\documentclass[letterpaper, 10 pt, conference]{ieeeconf}  

\IEEEoverridecommandlockouts                              
\usepackage{hyperref}
\usepackage{url}
\usepackage{graphicx} 
\usepackage{multirow}
\usepackage{caption}
\usepackage{subcaption}
\usepackage{amsmath}
\usepackage[utf8]{inputenc} 
\usepackage[T1]{fontenc}    
\usepackage{nameref}
\usepackage{caption} 
\usepackage{varioref}
\usepackage{hyperref}
\usepackage{cleveref}
\usepackage{float}
\usepackage{url}            
\usepackage{booktabs}       
\usepackage{amsfonts}       
\usepackage{nicefrac}       
\usepackage{microtype}      
\usepackage{lipsum}
\usepackage{dblfloatfix}
\usepackage{cite}
\usepackage{amsmath}
\usepackage{bbm}
\usepackage{autobreak}
\usepackage{xcolor}

\title{\LARGE \bf
Sim-to-Real Transfer for Robotic Manipulation with Tactile Sensory
}

\author{Zihan Ding$^{1}$, Ya-Yen Tsai$^{2}$, Wang Wei Lee$^{1}$, and Bidan Huang$^{1, \dag}$
\thanks{$\dag$ denotes the corresponding author.}
\thanks{$^{1}$Z. Ding, W.W. Lee, and B. Huang,  are with Tencent Robotics X, China  {\tt\footnotesize zhding@mail.ustc.edu.cn, \{wwlee, bidanhuang\}@tencent.com}}
\thanks{$^{2}$Y.-Y. Tsai is with the Hamlyn Centre for Robotic Surgery, Imperial College London, SW7 2AZ, London, UK {\tt\footnotesize y.tsai17@imperial.ac.uk}}
}

\begin{document}

\maketitle
\pagestyle{empty}

\begin{abstract}
Reinforcement Learning (RL) methods have been widely applied for robotic manipulations via sim-to-real transfer, typically with proprioceptive and visual information. However, the incorporation of tactile sensing into RL for contact-rich tasks lacks investigation. In this paper, we model a tactile sensor in simulation and study the effects of its feedback in RL-based robotic control via a zero-shot sim-to-real approach with domain randomization. We demonstrate that learning and controlling with feedback from tactile sensor arrays at the gripper, both in simulation and reality, can enhance grasping stability, which leads to a significant improvement in robotic manipulation performance for a door opening task. In real-world experiments, the door open angle was increased by 45\% on average for transferred policies with tactile sensing over those without it. 


\end{abstract}
\section{INTRODUCTION}


In human perception, tactile sensing is heavily relied on to acquire information about the local environment in a contact-rich task~\cite{dahiya2009tactile, lee2019neuro}. Tactile feedback can give access to supplementary information about the kinematics and mechanical properties of the environment that cannot be offered by other sensors such as vision. In a manipulation task, by making a contact with an object, tactile feedback provides additional properties about the object such as its pose in hand, surface texture and geometry, informing the user on the actions necessary to ensure safe handling during physical interaction with the object.

While the sense of touch is highly integrated in human perception, the same cannot be said about robotic manipulation. 
Compared to the sense of touch, other sensory information such as vision and force/torque is more commonly used in robotics~\cite{james2017transferring, akkaya2019solving, valassakis2020crossing, tsai2021droid}. 
Vision can provide global information of the environment such as spatial distances as well as local features like shapes and colors without the need of contact. In the event of visual occlusion, however, vision becomes insufficient. On the other hand, force/torque sensors are often mounted on the robot joints and are used to achieve proper compliance. Coupled with the appropriate force controller, joint based force/torque sensors enable safe robot execution in an uncertain and dynamic environment. Nonetheless, force/torque sensors cannot provide direct information about the contact area due to the intermedium of the kinematics chain and the entanglement of multiple contact forces. 
In a contact-rich task, tactile sensors can overcome the above deficiencies in vision and force/torque sensors by providing more local and direct contact details. 
\begin{figure}[t!]
	\centering\includegraphics[width=0.45\textwidth]{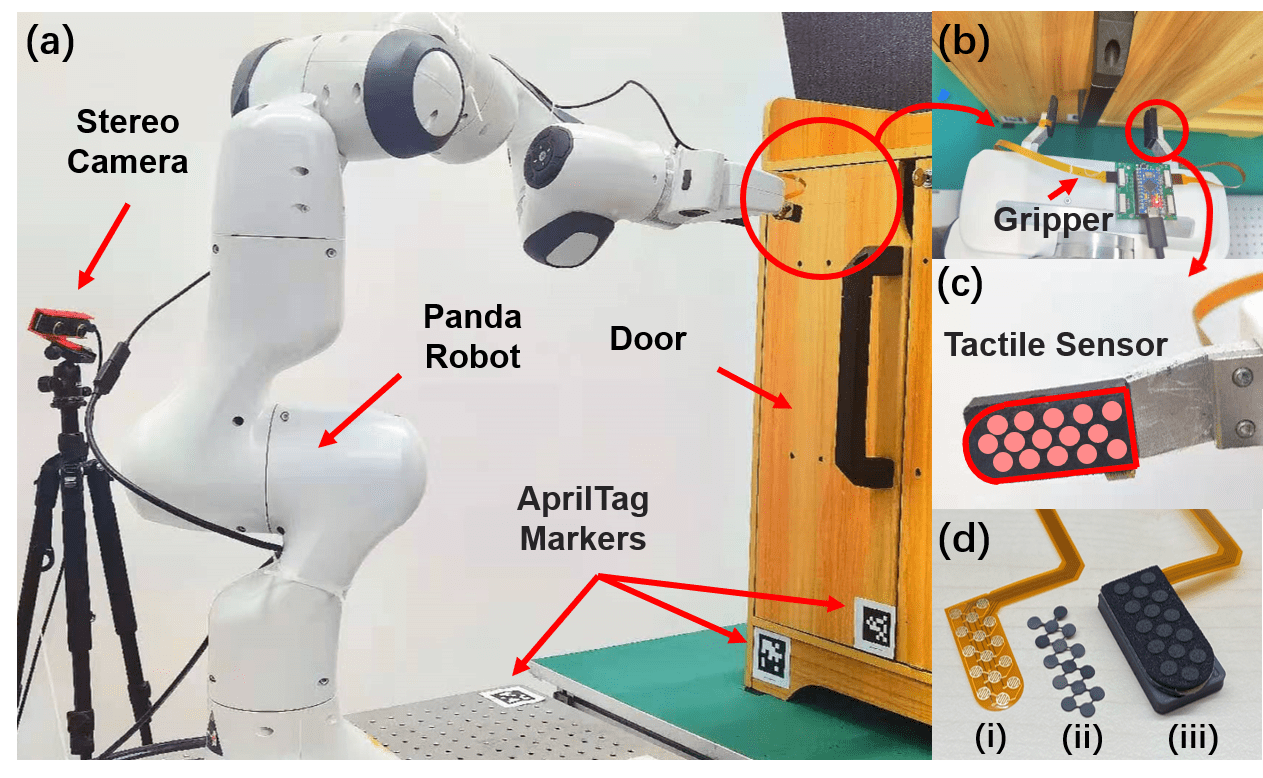}
	\caption{(a) and (b) show the experimental setup for robotic door opening in reality. (c) highlights the tactile sensor attached at the finger of the gripper. (d) shows the components of our tactile sensor array. From bottom to top, the sensor (iii) consists of inter-digitated electrodes fabricated on flexible PCB (i), piezoresistive polymer (ii) and the foam tape.}
	\label{fig:setup}
\end{figure}

There is an increasing interest in the usage of tactile sensors for robotic grasping and manipulation. However, the task of developing sensors that have high spatial and force sensitivity, precision and reliability remains a formidable challenge. In addition, reasoning about the state of the environment from tactile perception is often not straightforward due to spatially local information.
Many questions remain as to how tactile information should be incorporated into robotics control for the enhancement of manipulation performance.

\begin{figure*}[ht!]
	\centering
	\includegraphics[width=17 cm]{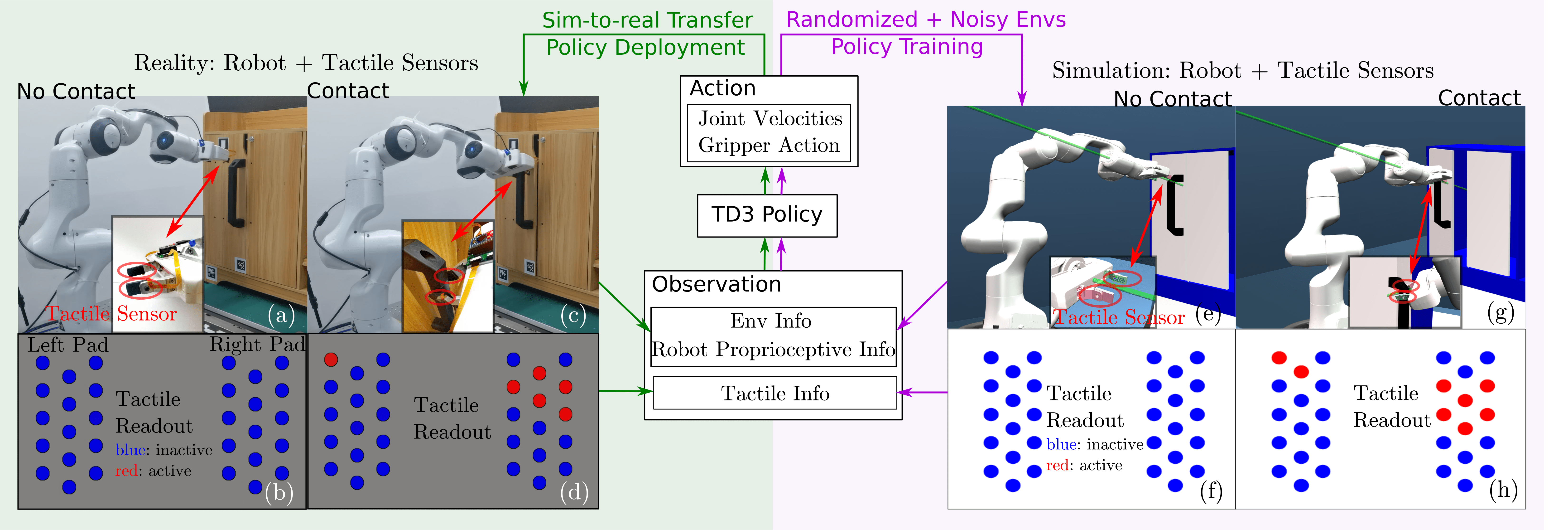}
	\caption{Overview of our method for policy training in simulation and sim-to-real transfer in reality using the Panda robot with tactile sensors for the door opening task. (a) and (c) show the real-world scenes for door opening using the robot with two tactile sensor arrays attached on the gripper pads, while (b) and (d) are corresponding tactile readouts for the contact and no-contact cases. (e) and (g) display the same settings in simulation as in (a) and (c), including simulated tactile sensors with force sensing. (f) and (h) are possible corresponding tactile readouts for (e) and (g). A deterministic policy is trained with TD3 algorithm in simulation with dynamics randomization and additional noises. After training, a policy is transferred into reality for real-robot control to finish the door opening task.  }
	\label{fig:overview}
\end{figure*}

Despite the recent success of Reinforcement Learning (RL) in robotics~\cite{akkaya2019solving, valassakis2020crossing, tsai2021droid, andrychowicz2020learning}, applications involving RL in real-robot grasping and manipulation using tactile sensing is still limited. One major reason is the prohibitive cost of conducting physical experiments on actual robots. Excessive experiments on physical robots will result in mechanical impairments as well as irreversible damage to fragile components such as tactile sensors. This is further exaggerated by the need for tactile sensors to make frequent contact with the environment in order to acquire data. To this end, training in simulation becomes a good alternative. Unfortunately, the simulation of tactile signals is not commonly provided on many simulation platforms. Furthermore, the policies trained in simulation may fail to work on the real robot due to the discrepancy between the simulation and the reality, \emph{i.e.}, the reality gap. Therefore, to incorporate tactile information in RL, it is crucial to 1) tackle the modeling of the tactile sensor in simulation and 2) cross the reality gap. 

In this paper, we constructed a model of our tactile sensor array that can be used in simulation. This model was then utilized for the training of a Twin Delayed Deep Deterministic Policy Gradient (TD3) RL algorithm~\cite{fujimoto2018addressing} within a simulated environment. The objective of the algorithm is to learn a policy for generating the respective joint velocities and gripper motions given the simulated environmental, proprioceptive and tactile feedback to accomplish a door opening task (Fig.~\ref{fig:overview}). The performance of the learned policy is then evaluated in reality. Our results show that even with the simplifications made to the modeling of tactile feedback in simulation, the policy that uses tactile feedback achieves better performance in the door opening task. The advantages were evident even when the policy was deployed without further adjustments after sim-to-real transfer. 

In brief, the contributions of this paper are as follows: we conducted studies to assess the usefulness of tactile feedback in a robotic manipulation task, and the feasibility of sim-to-real transfer for tactile-based policies. To facilitate the tactile-based RL policy learning and its sim-to-real transfer, a tactile sensor model was built and its response signals were processed to account for the inconsistent sensitivity and the discrepancy to the real signals. Finally, with domain randomization for policy training in simulation, we demonstrated a zero-shot sim-to-real transfer approach for improving robotic manipulation performance in real world using tactile sensing. 

\section{RELATED WORK}

Tactile information is essential in robotic grasping and manipulation as it gives access to local information about the contact region, such as the occurrence of collision between the robot and the environment, that is difficult to derive using other sensing modalities. A variety of tactile sensors have been developed to serve for different purposes, with types including optical-based~\cite{donlon2018gelslim, ward2018tactip}, capacitive pressure sensor array~\cite{lee2006flexible}, resistive sensor~\cite{russell1987compliant}, etc. GelSlim~\cite{donlon2018gelslim}, TacTip~\cite{ward2018tactip}, and BioTac~\cite{fishel2012sensing}
represent some of the more recent methods for tactile sensing in robotic research. GelSlim and TacTip are optical-based sensors that perceive the tactile information by observing the deformation of the contacting surface using a camera from within the sensor. BioTac detects the contact information with an array of electrodes placed on the surface of the sensor core. 

Previous works for tactile sensing in robotic literature usually involve tasks like object recognition~\cite{li2020skin, kaboli2019tactile,lee2017learning}, slip detection~\cite{james2018slip,reinecke2014experimental}, object shape reconstruction~\cite{meier2011probabilistic} and pose estimation~\cite{bauza2020tactile, bimbo2016hand, ding2020sim}, etc. These tasks typically leverage tactile information for inferring the state of the robot or the objects in direct contact with the robot. Another branch of works involves the additional usage of tactile sensing for assisting standard robot manipulation in tasks including object grasping~\cite{hogan2018tactile,li2014learning,chebotar2016self}, robotic stabilization~\cite{van2016stable}, edge following~\cite{lepora2019pixels}, etc. In ~\cite{li2020review}, the authors review the literature with tactile sensing for action selection on robots. These works show the significance and indispensability of the tactile sensor in grasping and other manipulation tasks. Our work falls into the second category of research, where we investigate the application of tactile information in robotic manipulation.

Most of the above works were developed directly on a physical tactile sensor without leveraging simulation. In reality, data collection and other extensive tests could accelerate the impairment of tactile sensors and thus limiting its applications in many robotic manipulation learning tasks. Custom made tactile sensor simulations have been reported~\cite{ding2020sim, narang2020interpreting, wang2020tacto}, but their performance in sim-to-real policy transfer for robotic manipulations have not been investigated. Accurately modeling a tactile sensor with little or no real-world data still pose extreme difficulty. 
A few studies have reported promising results involving tactile sensing for policy learning in simulation~\cite{huang2019learning}. Wu et al.~\cite{wu2019mat} used the embedded tactile sensor on the Barrett hand and developed a closed-loop RL algorithm to optimize a robotic grasping strategy. The experiment results show that utilizing tactile information could improve the success rate of grasping. Our approach draws some similarities to their work in relying simulation to train a RL policy and incorporating the tactile information into our observation. Different from their work, however, our focus places on a manipulation task, i.e. opening a door, and the sim-to-real transfer performance with tactile sensing. Here, we present a successful zero-shot transfer for polices which are learned from scratch in simulation, with domain randomization techniques applied. 

\section{PRELIMINARIES}
\label{sec:pre}
\subsection{RL Formulation}
To apply the standard RL methods, the robot learning environment is formulated as a Markov Decision Process (MDP), which can be represented as $(\mathcal{S}, \mathcal{A}, R, \mathcal{T}, \gamma)$. $\mathcal{S}$ and $\mathcal{A}$ are the state space and the action space, and $R$ is a reward function $R(s,a)$: $\mathcal{S}\times \mathcal{A}\rightarrow \mathbb{R}$ for current state $s\in\mathcal{S}$ and action $a\in\mathcal{A}$. 
The state-transition probability from current state and action to a next state $s^\prime\in\mathcal{S}$ is defined by $\mathcal{T}(s^\prime|s,a)$: $\mathcal{S}\times\mathcal{A}\times\mathcal{S}\rightarrow \mathbb{R}_+$. $\gamma\in(0,1)$ is a reward discount factor. For cases with deterministic policy, the agent will follow the policy $\pi$: $\mathcal{S}\rightarrow\mathcal{A}$, to generate its action $a=\pi(s)$ given the current state $s$, and receive a reward $r=R(s,a)$. The goal of an RL algorithm is optimizing the policy to maximize the agent's expected cumulative rewards: $\mathbb{E}[\sum_t \gamma^t r_t]$. 

\subsection{Twin Delayed Deep Deterministic Policy Gradient}
TD3~\cite{fujimoto2018addressing} is a modified version of Deep Deterministic Policy Gradient (DDPG)~\cite{lillicrap2015continuous} to empirically address the function approximation error for solving continuous control tasks, by incorporating additional techniques like clipped double-Q learning and delayed policy update. We denote the deterministic policy as $\pi_\theta(\cdot)$ with neural network parameters $\theta$. The policy is trained with an objective: $\max_\theta\mathcal{J}_\theta=\max_\theta\mathbb{E}_{\pi_\theta}[\sum_t \gamma^{t}r_t]$, which is often approximated as $\max_\theta\mathbb{E}_{s\sim\mathcal{D}}[Q_{\phi_1}(s, \pi_\theta(s))]$ in practice. There are two Q-networks (indexed by j=1,2) in TD3 parameterized by $\phi_j$, with the learning objective:
\begin{align}
    L(\phi_j, \mathcal{D}) &= \mathbb{E}_{(s,a,s^\prime,r,d)\sim\mathcal{D}}[(Q_{\phi_j}(s, a)-Q^T)^2], j=1,2\\
    Q^T&= r(s,a)+\gamma(1-d)\min_{i=1,2}Q_{\phi_i}(s^\prime,\pi^T_\theta(s^\prime)+\epsilon)
\end{align}
where $d$ is a boolean value to indicate trajectory finishing, $\pi^T_\theta$ is the target policy and $\epsilon$ is a noise term for target policy smoothing, and $\mathcal{D}$ is the off-policy dataset for training. 

\section{METHODOLOGY}
Our proposed method involves the modeling of the tactile sensor array in simulation and the sim-to-real transfer scheme to verify tactile sensing for assisting robot manipulations. For the tactile sensor, we describe the details in design and simulated modeling of our tactile sensor arrays in Section~\ref{subsec:tactile} and \ref{subsec:tactile_model}. Our sensors provide more consistent response to forces applied and help alleviate the difficulties in simulated modeling due to its inconspicuous deformation in collisions. For the sim-to-real policy transfer, an overview is depicted in Fig.~\ref{fig:overview}. Descriptions about the RL task settings are provided in Section~\ref{subsec:training}, and the sim-to-real transfer methods applied in our scheme is described in Section~\ref{subsec:sim2real}. 


\subsection{Tactile Sensor Array}
\label{subsec:tactile}
\begin{figure}[b]
    \centering
    \includegraphics[width=0.35\textwidth]{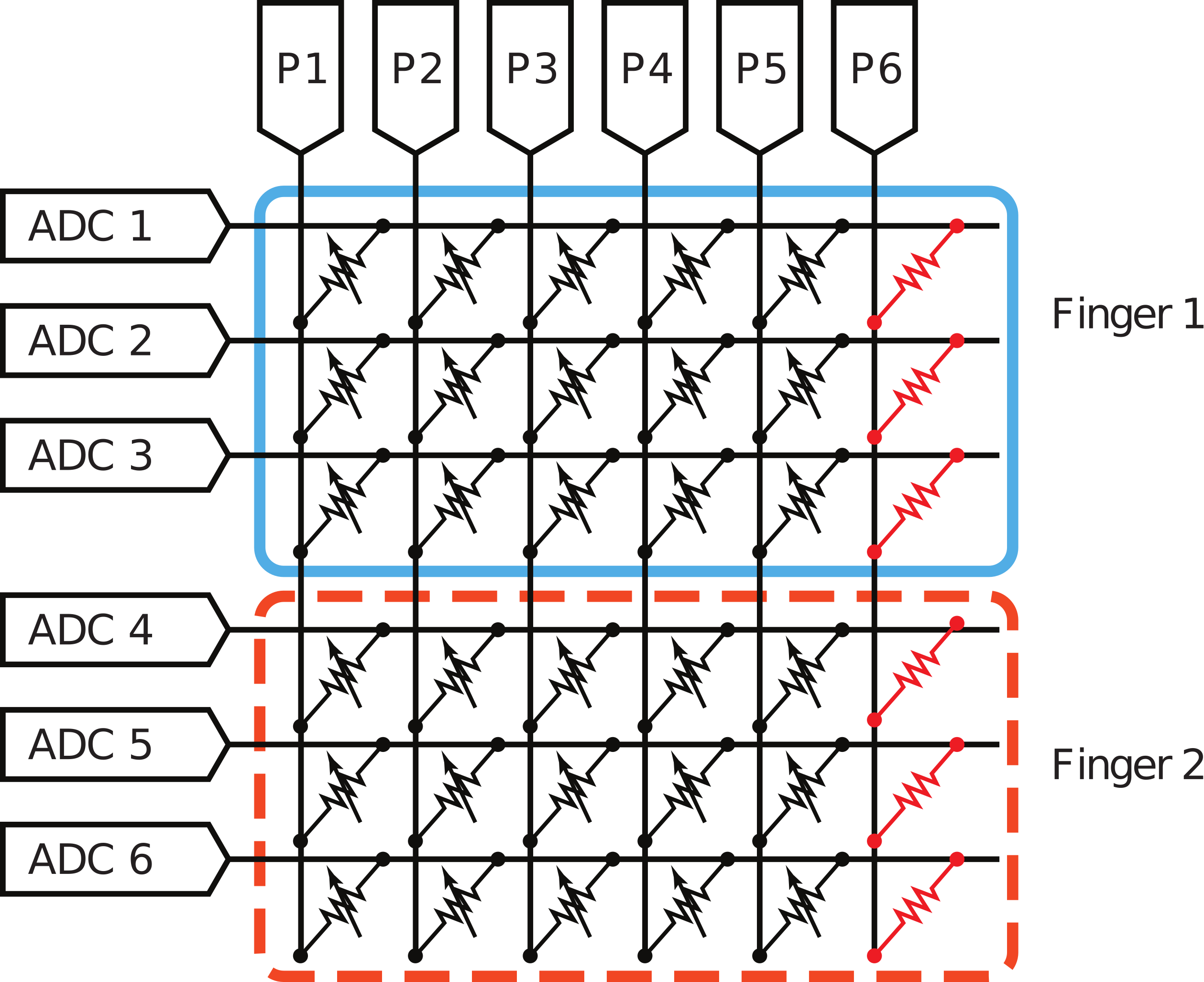}
    \caption{The schematic circuitry for the tactile sensor array. Each sensor element is represented as a variable resistor. Pins P1-6 are digital pins that drive the columns at either 5V or ground. An additional known resistor per row (red) is placed in the circuit for computational purposes.}
    \label{fig:sensor_schematics}
\end{figure}

To facilitate this study, resistive tactile sensor arrays were developed in-house. The sensing elements in Fig.~\ref{fig:setup}~(c)(d) consist of inter-digitated electrodes fabricated using flexible printed circuit boards. Carbon impregnated polymer sheets (Velostat, 3M) were used for transduction. To improve compliance and concentrate contact forces onto each element, foam bumps were fabricated over each electrode by laser etching of EVA foam tape (3M 9080). The foam tape also served to hold the transducer sheet in place. The system has a total of 30 sensor elements (arranged as a 6$\times$5 matrix) which were evenly distributed between 2 pads attached onto both sides of the robotic gripper.

The equivalent electrical circuit of the sensor array is shown in Fig.~\ref{fig:sensor_schematics}, where each sensor element is modeled as a variable resistor. Increases in pressure at each element will result in a reduction in resistance. To compute the resistance distribution in the array, we used the resistance-matrix approach adapted from \cite{Shu2014}. A column is first sampled by driving it to 5V (digital high) while connecting the other columns to ground. The rows are then read using an ADC. After scanning the entire row, the column is driven to ground while the next column is set to 5V. The process is repeated until the entire array is scanned. Cross-talk in the circuitry can then be canceled using the computation described in \cite{Shu2014}. The process is performed using a microcontroller (ATmega32u4, Atmel) and transmitted to the PC via USB, at a rate of about 200 frames per second.

\subsection{Tactile Sensor: Modeling in Simulation}
\label{subsec:tactile_model}

MuJoCo physics engine ~\cite{todorov2012mujoco} is leveraged for simulation. In simulation, the geometric model of a tactile sensor is measured explicitly. On the real sensor, each sensor unit (corresponding to a single bump) responds to a perpendicular pressing force in the form of an electrical signal and their relationship follows a non-linear but positive correlation. Considering the factors such as highly non-linear sensor feedback, material deformation of the real tactile sensor and the limitation of the simulator, explicit modeling of the real electrical response is unrealistic at present. Instead, an approximated model is built by placing a force sensor beneath each unit. Each unit is modeled as an independent cylinder body attached on the pad body. Although the force sensor in MuJoCo can provide a 3-dimensional continuous readout, only the perpendicular force is considered to simulate the electrical signal obtained from the real tactile device. A scaling is done to match the magnitude between the force and the electrical response. To account for the aforementioned modeling difficulties and to minimize the potential reality gap, we further process the force and the electrical responses in simulation and reality to binary values for robot learning. The threshold is set heuristically to convert the continuous value to binary. Note that, in MuJoCo, we do not use a touch sensor to simulate the electrical signal due to its lower sensitivity in simulation from our empirical comparisons.



\subsection{RL Task Settings}
\label{subsec:training}

A door opening task is chosen to evaluate the practicability and the usefulness of tactile sensor in manipulation problem. This task involves bringing the robot to interact with the environment from a non-contact state to a contact state and maintaining it, and handling the dynamics of the robot joints and the door hinge. It consists of three phases: door approaching, knob grasping and door opening. The knob grasping and door opening actions require a direct contact with the door knob, which will mutually affect the states of both the robot and the door. A good grasp of the door knob is essential and critical to the later door opening process, while a bad grasping pose can lead to door knob slipping from the gripper. Hence, the task is carried out with tactile sensing to verify if the tactile information could improve the grasping performance and thus the door opening task. 

In this setting, we incorporate the tactile information as observation in a typical RL framework for robot learning. The details of the framework, including the observation space, action space, reward shaping and policy learning settings are described in the following paragraphs. 

\paragraph{Observation Space} In the absence of tactile information, the observation of the robot agent in the door-opening environment contains a total of 25 dimensions, which involves the robot proprioceptive information including the end-effector position, end-effector velocity, joint positions, joint velocities and gripper width, as well as environment information, i.e. the position of door knob relative to the end-effector and the angle of the door hinge. In our primary tests, the robot is already capable of approaching, grasping and opening the door with above single-step observation, so we do not add historical observations as inputs to the control policy. By incorporating tactile information in the framework, we include additional 30 binary tactile values, $\{\hat{c_i}|i=1,2,...,30\}$ in the observation space. Each of these values is converted from the tactile electrical signal into binary value using a threshold $\kappa$ as follows:
\begin{equation}
    \hat{c_i}=\begin{cases} 1, \text{ if } c_i>\kappa , i=1,2,...,30\\ 0, \text{ otherwise}.
    \end{cases}
\end{equation}

\paragraph{Action Space} The action space has 8 DoF. The first 7 DoF are the desired joint velocity to control a 7 DoF robot, and the last DoF commands the gripper motion. During training, these values are normalized.


\paragraph{Reward Shaping}
We provide a shaped reward function $R$ at each timestep in simulation for improving the learning efficiency, which is similar to~\cite{tsai2021droid} and defined as follows:
\begin{align}
    R &= \omega_{door} \cdot r_{door} + \omega_{dist} \cdot r_{dist} + \omega_{ori} \cdot r_{ori} \notag\\
    & + \omega_{grasp} \cdot r_{grasp} + \omega_{tactile} \cdot r_{tactile}
    \label{equ:reward}
\end{align}
The above function involves five terms, $r_{door}$, $r_{dist}$, $r_{ori}$, $r_{grasp}$, and $r_{tactile}$, which are defined as following:
\begin{align}
    r_{door} &= \begin{cases} \alpha, \text{if } \mathbf{1}_{\text{grasp}}, \\
    0, \text{otherwise},
    \end{cases} \\[1em]

    r_{dist} &= -1-tanh(||\mathbf{x}_{knob} - \mathbf{x}_{gripper} ||_2), \\[1em]
    
    r_{ori} &= -1-tanh(||\Omega_t - \Omega_g ||_2), \\[1em]
    
    r_{grasp} &= \begin{cases} 1, \text{if } \mathbf{1}_{\text{grasp}}, \\
    0, \text{otherwise},
    \end{cases} \\[1em] 
    r_{tactile} &= \begin{cases} ||\boldsymbol{\hat{\mathbf{c}}}||_1, \text{if }\alpha>\alpha_0 \text{ and } \mathbf{1}_{\text{grasp}}, \\
    0, \text{otherwise},
    \end{cases} 
    \label{equ:reward_tactile}
\end{align}
In the above equations, $\alpha$ is the hinge angle of the door in a range $[0^\circ, 90^\circ]$ and $\alpha_0 = 1.15^\circ$ is an empirical threshold value. $\mathbf{1}_{\text{grasp}}$ indicates the grasping state with value 1 when both gripper fingers are in contact with the knob, otherwise 0. $\mathbf{x}_{knob}$ and $\mathbf{x}_{gripper}$ are 3-dimensional positions of the knob center and the center of the gripper respectively. For the orientational reward $r_{ori}$, both the 3-dimensional orientation vectors of the gripper $\boldsymbol{\theta}_g$ and its target (\emph{i.e.}, the orthogonal direction of the initial door surface) $\boldsymbol{\theta}_t$ are sine-cosine encoded as used in \cite{valassakis2020crossing}: $\Omega_t=(sin(\boldsymbol{\theta}_t),cos(\boldsymbol{\theta}_t)),\Omega_g=(sin(\boldsymbol{\theta}_g),cos(\boldsymbol{\theta}_g))$. $\boldsymbol{\hat{\mathbf{c}}}$ consists of 30 binary values corresponding to the activation state of each unit in tactile sensors at current timestep. The value $-1$ in the $r_{dist}$ and $r_{ori}$ terms ensures their non-positive values. $\omega_{door}, \omega_{dist}, \omega_{ori}, \omega_{grasp}, \omega_{tactile}$ are normalizing coefficients of different reward terms, with values 5.0, 0.4, 0.05, 0.1, and 0.01. 

\paragraph{Policy Learning} The TD3 algorithm is employed for learning the robotic control policy in the door opening task. A distributed training scheme is applied for the policy learning process with multiprocessing across multiple GPUs and CPU cores. During training, the exploration noise for actions is decayed over time to facilitate stabilization.

\subsection{Sim-to-real Transfer}
\label{subsec:sim2real}
Domain randomization technique is adopted in RL training to enhance the robustness in sim-to-real transfer. 
Due to the randomization of dynamics parameters, the transition function $\mathcal{T}$ of the simulated system becomes $\mathcal{T}(s^\prime|s,a;\xi)$, which is now dependent on the parameters, $\xi$, sampled from distribution $p(\xi)$. The corresponding RL objective is modified from the standard RL objective defined in Section~\ref{sec:pre} to the following:
\begin{equation}
    \max_\theta\mathbb{E}_{\xi\sim p(\xi)}[\mathbb{E}_{s\sim \hat{\mathcal{D}}}[Q(s, \pi_\theta(s))]]
\end{equation}

where $\hat{\mathcal{D}}$ is a dataset containing data samples in a format of $({\hat{s}+\epsilon_s}, a+\epsilon_a, \hat{s^\prime}+\epsilon_{s^\prime}, r, d)$, and $\hat{s^\prime}$ is from $\mathcal{T}(\hat{s^\prime}|\hat{s},a;\xi)$ instead of $\mathcal{T}(s^\prime|s,a)$ due to the randomized dynamics. $\epsilon_{s}, \epsilon_{s^\prime}$ and $\epsilon_a$ are observation noise and action noise sampled from given distributions.


\paragraph{Dynamics Randomization}
To mitigate the reality gap in the door-opening task, we randomize the critical parameters in Table~\ref{tab:rand1}, that cannot be measured accurately in practice, within manually determined ranges. We select the randomized ranges such that it covers the data distributions of the reality. During policy training, each parameter is sampled from a uniform distribution to create a unique environment setting in simulation. Note that we only randomize the environment parameters excluding the robot's because they are subjected to changes in different settings. The parameters of the robot, including the mass of links, joint damping, proportional gains for joint control and others, are fixed and explicitly identified through trajectory alignment process. This is done through matching the joint position trajectories in simulation and reality before all experiments. 

\begin{table}[htbp]
\begin{center}
\captionof{table}{Randomized dynamics parameters}
\begin{tabular}{ |c|c|c| } 
\hline
\textbf{Parameter} & \textbf{Range} \\
\hline
Knob Friction & [0.8, 1.0] \\ \hline
Door Hinge Stiffness & [0.1, 0.8] \\ \hline
Door Hinge Damping & [0.1, 0.3] \\ \hline
Door Hinge Friction Loss  & [0.0, 1.0] \\ \hline
Door Mass & [50.0, 150.0] \\ \hline 
Knob Mass & [2.0, 10.0]  \\

\hline
\end{tabular}
\label{tab:rand1}
\end{center}
\end{table}

\paragraph{Additional Noise/Delay}
In addition to randomizing the door environment, we also implement in our method some additional noise and delay factors for handling the unmodeled effects~\cite{akkaya2019solving} as well as achieving more generalized policies. These unmodeled effects can be caused by inaccurate localization of objects with the visual sensor, as well as jerking or gear backslash in the robot, etc. Therefore, we add observation noise $\epsilon_o$ and action noise $\epsilon_a$ into observations and actions of the robot during training. Both noises are sampled from uniform distributions with ranges displayed in Table~\ref{tab:rand2}. For the observation, only the 30-dimensional binary tactile readout is not affected by this noise; while for the action noise, the gripper control is left out. We also allow in simulation to have a single-step observation delay to happen in training, for all observation dimensions. All noise and delay are added in a timestep scale (\emph{i.e.}, one set of values for each single timestep). The last two parameters are the position offsets of the table in X-and Y-axis, which is found to be useful to account for the small offsets in the distance between the position of the table (and door) and the robot in reality. The position offset is applied in an episode scale (\emph{i.e.}, one pair of values for each episode). 
\begin{table}[htbp]
\begin{center}
\captionof{table}{Noise/delay parameters}
\begin{tabular}{ |p{1.8cm}|c|p{2.5cm}|c| } 
\hline
\textbf{Parameter} & \textbf{Range} & \textbf{Applied Variables} &\textbf{Scale}\\
\hline
Observation Noise & [-0.002, 0.002] & all obs but not tactile & timestep \\ \hline
Observation Delay & $\{0, 1\}$ & all obs & timestep \\ \hline
Action Noise & [-0.01, 0.01] & all action but not gripper & timestep\\ \hline
Table Position Offset (X-axis) & [-0.05, 0.05] & table position X & episode\\ \hline 
Table Position Offset (Y-axis) & [-0.05, 0.05] & table position Y & episode \\
\hline
\end{tabular}
\label{tab:rand2}
\end{center}
\end{table}

\paragraph{Tactile Signal Randomization}
As described in Sec.~\ref{subsec:training}, we apply the binarized tactile signals (30 dimensions) in observation during robot movements. In order to mitigate the reality gap for tactile sensors, we use a method with ``binary flipping" (0 to 1, 1 to 0) to process the tactile readouts in training. More concretely, at each timestep, every binary value of a tactile sensor readout is going to be flipped with a certain probability $p_{\text{flip}}$. In practice, we use $p_{\text{flip}}=0.005$.

\section{EXPERIMENT}

In our experiment, a door opening task was served for the purpose of determining whether the tactile sensor could enhance the manipulation performance in a contact-rich environment. This task involves complex dynamics of the robot and the environment, which has been studied in many previous works including DoorGym~\cite{urakami2019doorgym}. In this work, we replaced the hook end-effector in the DoorGym setting with a two-finger gripper for more general purposes. This also introduces more challenging contact and manipulation uncertainties, which we attempted to tackle with the tactile sensors. 

To cross the reality gap, we first made a detailed analysis to understand the discrepancy between the simulated and the real tactile sensor, and implemented the calibration method as in Section~\ref{subsec:tactile_compare}.
In the second experiment, we deployed the RL policies with and without tactile information in the real world for direct comparisons, detailed in Sections~\ref{subsec:door_setup} and \ref{subsec:door_result}.



\subsection{Tactile Sensor: Sim-to-Real}
\label{subsec:tactile_compare}
\begin{figure}[t]
	\centering\includegraphics[width=0.45\textwidth]{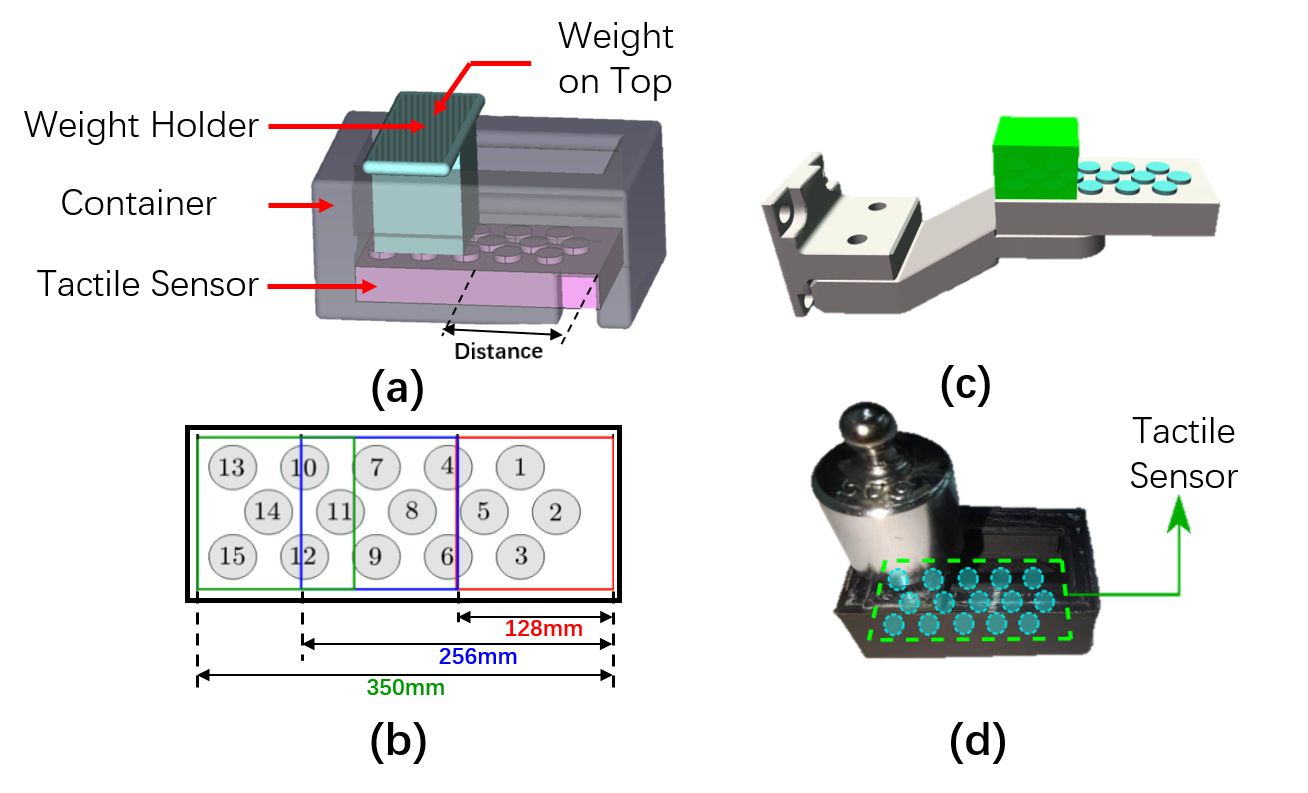}
	\caption{Experimental settings for comparing the readout of a single tactile sensor array in simulation and reality. (a) is the assembly of the testing device. (b) shows the square regions at different distances (128mm, 256mm, 350mm) pressed in the tests, and the numbering of the 15 sensor units. (c) and (d) are testing case with distance=350mm for simulation and reality respectively. }
	
	\label{fig:tactile_test}
\end{figure}

\begin{figure*}[!h]
	\centering\includegraphics[width=0.95\textwidth]{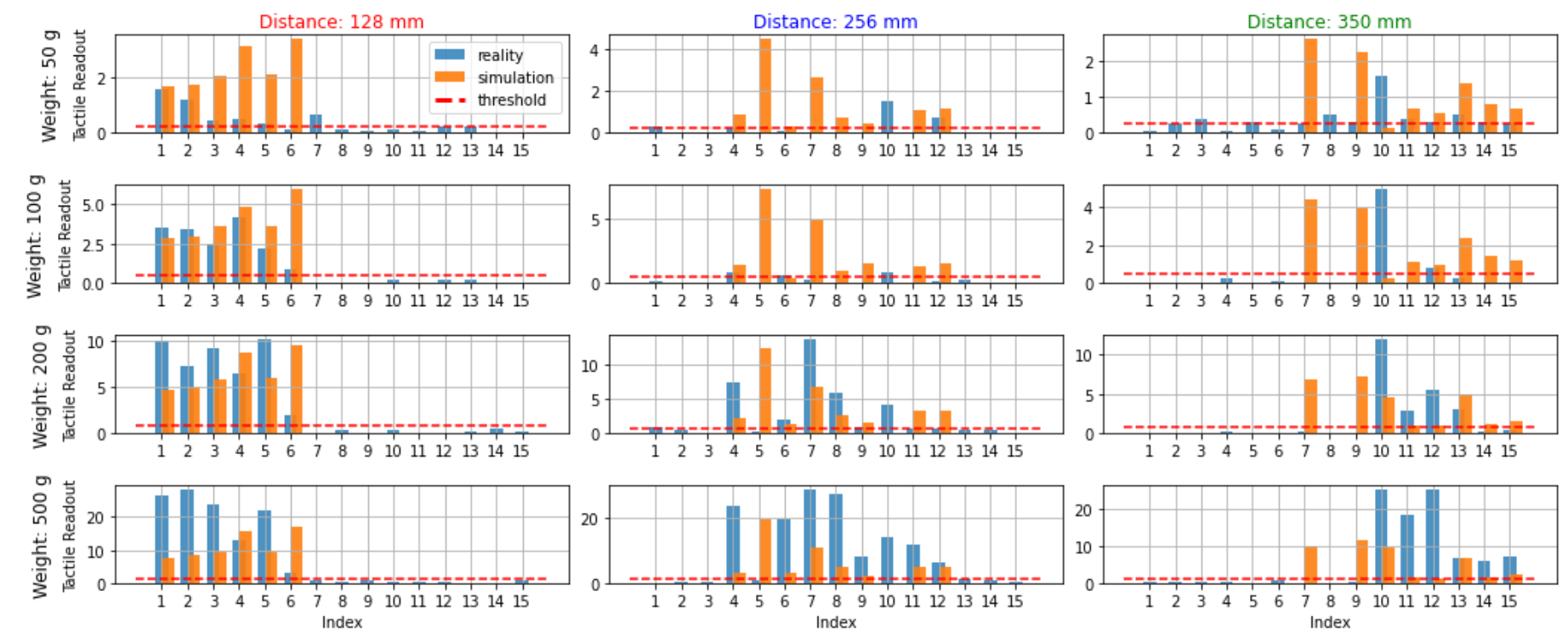}
	\caption{Comparison of tactile array (15 units) signals in simulation and reality. The horizontal axis is the unit index (1-15) on the sensor. The sensor readout in reality is multiplied by a factor of $5\times10^5$ for scaling. 
	The red dashed lines indicate the threshold $\kappa$ for the binarization of tactile signals, with values of 0.25, 0.5, 0.75, 1.25 for weight 50g, 100g, 200g and 500g.}
	\label{fig:tactile_compare}
\end{figure*}

To achieve the sim-to-real transfer for the simulated tactile sensors, we aligned the tactile signals between simulation and reality with a calibration process. 
Fig.~\ref{fig:tactile_test} shows the experimental settings.  Fig.~\ref{fig:tactile_test}~(a) depicts the devices assembly, including a weight holder (cyan), a container (grey) a container for positioning the sensor and the weights, and a tactile sensor array (pink) with 15 units at the bottom of the container, being pressed by the bottom surface of the weight holder. Different weights can be put on top of the weight holder to generate even pressure on the sensor with desired forces. Each of the device has a simulated counterpart. Fig.~\ref{fig:tactile_test}~(b) pictures the numbering of each sensor unit, as well as the square regions with different distances (128mm, 256mm, 350mm) along the sensor. Fig.~\ref{fig:tactile_test}~(c) and (d) illustrate the calibration process in simulation and reality, from which we studied how the sensor units were activated under different force distributions. As shown in Fig.~\ref{fig:tactile_test}~(c), those devices were simplified to a single rectangular block with different weights (equivalent to weight plus weight holder in reality).




During the experiment, we tested three pressing positions at 128mm, 256mm, and 350mm along the tactile sensor to activate different sensor units using four weights, 50g, 100g, 200g, and 500g.  Fig.~\ref{fig:tactile_compare} shows the real and simulated signals (normalized) of a tactile sensor under these different force distributions. From this figure we can see that in most of cases the real and simulated tactile sensors were activated at the same time, though their values do not show a simple linear correlation. This is expected as the real sensor is covered by a soft foam while the simulated sensor is modeled as a series of rigid bodies. According to these results, we determined the threshold $\kappa$ of each sensor unit and use them as binary observations of contact in the policy learning. 


\subsection{Door Opening Setup}
\label{subsec:door_setup}
For the door opening task, the experiments were conducted with a 7 DoF Franka Emika Panda robot as shown in the Fig.~\ref{fig:setup}. During the interaction with the door knob, the Panda robot was commanded at 20Hz through the joint velocity controller and was controlled at 1kHz. To ensure smooth motion, a polynomial interpolation was done between consecutive commands. An equipped gripper was used for grasping. Visual information was provided through a binocular camera, ZED Mini from Stereolabs. The poses of the cabinet and the door were detected through AprilTag markers~\cite{olson2011apriltag}. To perceive tactile information, the original finger pads on the gripper were replaced by a pair of in-house resistive tactile sensors as detailed in Section~\ref{subsec:tactile}. A simulation environment was developed in MuJoCo to mirror this real setting, including the tactile sensors on the robot's fingertips. The relative poses between objects, including the door hinge angle, was directly obtained through simulation and the tactile feedback was perceived by reading the force sensor placed on the gripper as described in Section~\ref{subsec:tactile_model}. The robot learned policy through a model-free RL approach via simulation. Then, the policy was deployed to the real environment to carry out the door opening task. 






\subsection{Door Opening w/o and w/ Tactile}
\label{subsec:door_result}
\begin{figure}[htbp]
	\centering\includegraphics[width=0.4\textwidth]{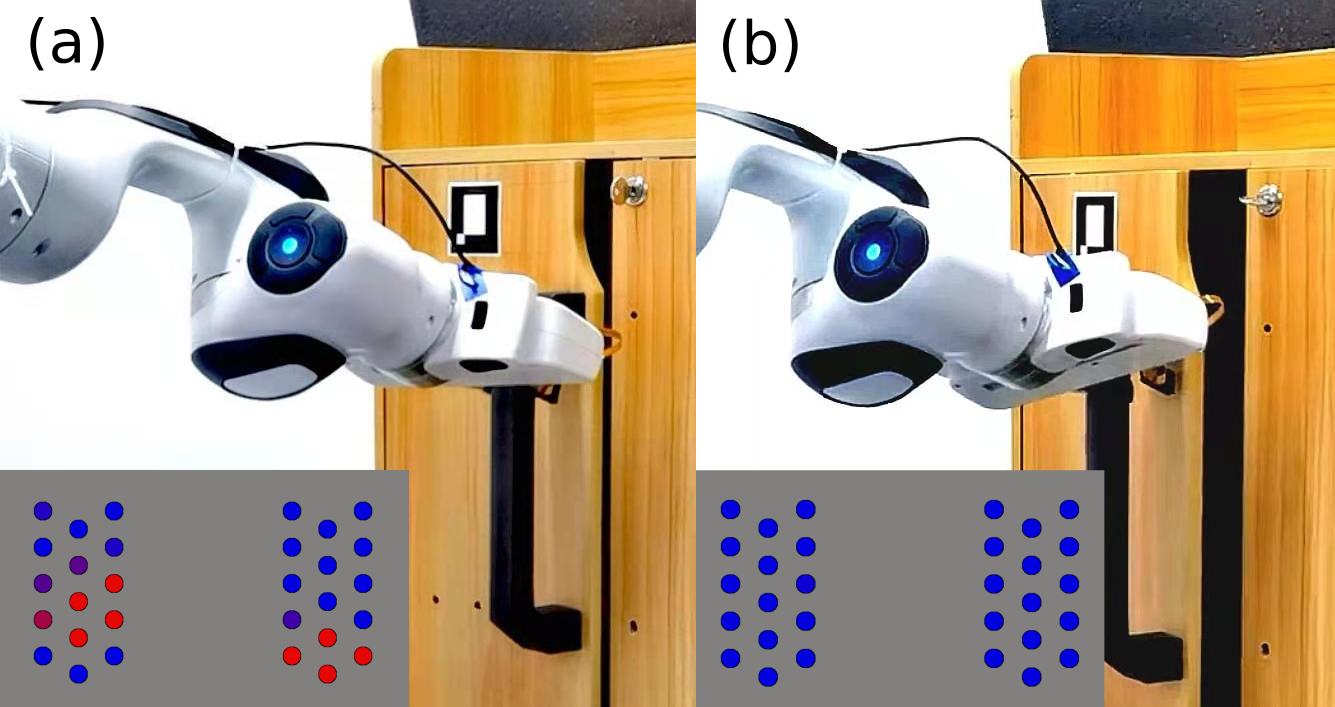}
	\caption{Comparison of the door opening poses in reality for the robot with (a) and without (b) tactile sensing.}
	\label{fig:real_compare}
\end{figure}

In simulation, we trained three policies for either case with or without tactile sensing using TD3 algorithm on different random seeds, and kept all the other settings unchanged. Each policy was learned with distributed training scheme running over a total of 25000 episodes using 5 processes. Each episode took a maximum of 1000 steps to carry out the task. Dynamics randomization and additional noise/delay described in Section~\ref{subsec:sim2real} were employed during training to improve the robustness in sim-to-real transfer. For both simulation and reality, these 3 policies were evaluated for 10 times each, to acquire the average performances. 

\begin{figure}[htbp]
	\centering\includegraphics[width=0.4\textwidth]{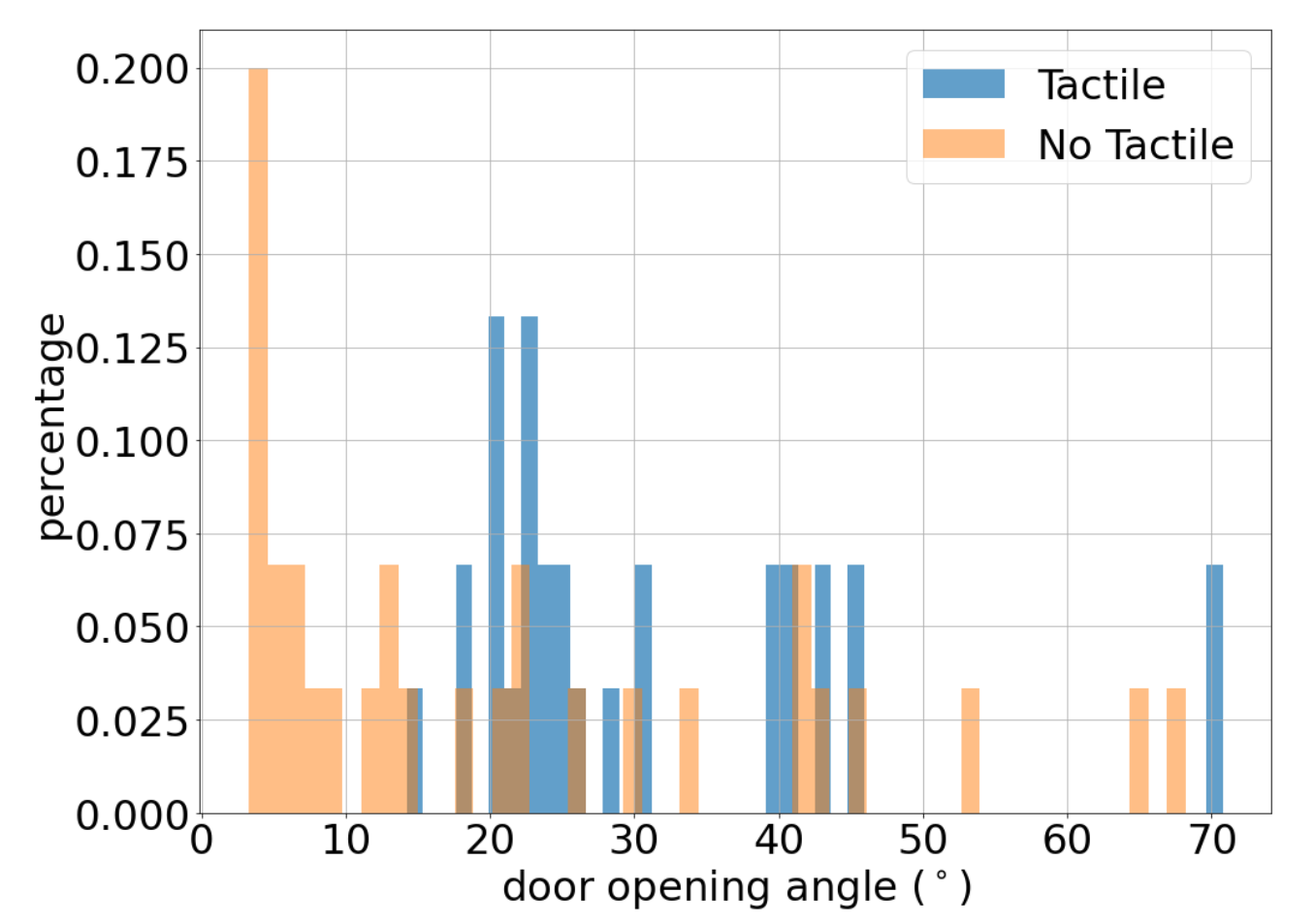}
	\caption{Comparison of door opening angles in reality for the robot with and without tactile sensing. The percentage values were calculated based on 30 trials in reality for each setting.}
	\label{fig:compare}
\end{figure}

Fig.~\ref{fig:real_compare} shows the door opening strategies from policies with and without tactile sensing. The policy leveraging tactile information appeared to demonstrate a better grasping pose as it positioned the gripper to grasp the sides of the door knob with many tactile units activated, which implies large contact region. Large contact area minimized the chance of slipping and hence lead to a better grasp stability and more consistent door opening result shown in Fig.~\ref{fig:compare}. 


Quantitatively, from Table~\ref{tab:result}, we could see that by incorporating the tactile feedback, the agent was able to achieve a larger mean of door opening angle in both simulation and reality, compared to the results from the policies that didn't use such information. It also resulted in greater minimal and maximal door-opening angles and a significant decrease in the number of steps taken to open the door, from an average of 275.6 to 176.3. This indicates an increase in task performance and a faster and smoother control of the robot can be achieved. The `Steps' column in the table indicates the total steps taken to achieve the maximal angles. In reality, the robot took fewer steps than in simulation because the maximal angles achievable were usually smaller. By inspecting the standard deviation results from both simulation and reality, we could further deduce that with tactile, the learned policies were able to achieve a more consistent performance. 

We additionally presented the episode reward for further comparison. A direct comparison may seem to be unfair as the reward function for no-tactile cases had one less tactile reward term as Eq.~\eqref{equ:reward_tactile}. Therefore, we could consider the maximum episode reward that could be achieved by policies without tactile, by adding the maximum tactile reward term, ${timesteps}^{max}\cdot\omega_{tactile}\cdot r^{max}_{tactile}=300$, to the average episode reward listed in the table. However, even with this included, the episode reward obtained with tactile sensor still outperformed, which demonstrated the usefulness of tactile sensors in solving the door opening task.

Based on the experimental results obtained, we verified that the ``binarized'' tactile information is advantageous in getting a better grasping pose, which is of extreme critical in achieving later better door opening performance and consistency.

\begin{table}
\centering
\begin{tabular}{ |p{4mm}|p{12mm}|p{10mm}|p{12mm}|p{12mm}|p{12mm}| } 
 \hline
 &  & Door Angle ($^\circ$) & Angle Min/Max ($^\circ$) & Steps & Reward \\ \hline 
\multirow{ 2}{*}{Sim} &  w/ tactile & $\mathbf{41.8}\pm\mathbf{15.7}$ & 0.6 / 90.0  & $720.5\pm234.4$ & $\mathbf{2435.6}\pm\mathbf{1024.6}$\\  \cline{2-6}
& w/o tactile & $34.6\pm20.3$  & 1.3 / 90.0 & $584.4\pm273.4$ & $1881.8\pm1363.3$ \\ \hline
\multirow{ 2}{*}{Real} &  w/ tactile & $\mathbf{31.2}\pm\mathbf{14.0}$ & \textbf{14.2 / 70.8} & $\mathbf{176.3}\pm\mathbf{25.8}$  & -  \\ \cline{2-6}
&w/o tactile  & $21.5\pm18.9$ & 3.3 / 68.2 & $275.6\pm167.2$ & -\\
 \hline
\end{tabular}
\caption{Comparison of policy performance.}
\label{tab:result}
\end{table}
\section{CONCLUSIONS}
Enriching the robot sensory inputs for better optimizing the operations of robot is a long-term challenge for the community. Learning-based methods provide the capability of handling with high-dimensional sensor modalities involving vision and touch. Our investigation in this paper testified the feasibility of applying approximately modeled tactile sensor arrays to assist robot learning in simulation and transferred policies for robot manipulations with tactile sensing in reality, which might constitute a necessary component in advanced robotic control for solving complex contact-rich tasks. The overall framework is not only feasible for our in-house tactile sensor arrays and the chosen door opening task, but also for other tactile sensors with relatively low granularity and inconspicuous deformation, on general contact-rich tasks. Further verification of our proposed framework on different sensors and tasks will be studied in our future work. By leveraging the binary readouts of tactile sensors as additional input, the RL agent could perform better grasping and opening the door in both simulation and reality. In future works, we will also attempt at building a better modeling of similar tactile sensors in simulation and directly leveraging their continuous readouts for policy enhancement.

\bibliography{ref}
\bibliographystyle{IEEEtran}

\end{document}